\documentclass[runningheads]{llncs}

\usepackage{eccv}

\usepackage{eccvabbrv}

\usepackage{graphicx}
\usepackage{booktabs}
\usepackage{color}
\usepackage{multirow}
\usepackage{multicol}

\usepackage[accsupp]{axessibility}  

\usepackage{hyperref}

\usepackage{orcidlink}

\begin{document}

\title{Strictly-ID-Preserved and Controllable Accessory Advertising Image Generation} 

\titlerunning{Abbreviated paper title}

\author{Youze Xue\inst{1} \and
Binghui Chen\inst{2} \and
Yifeng Geng\inst{2} \and Xuansong Xie\inst{2} \and Jiansheng Chen\inst{3} \and Hongbing Ma\inst{1}}

\authorrunning{F.~Author et al.}

\institute{Tsinghua University \and
Institute for Intelligent Computing, Alibaba Group\\
 \and
University of Science and Technology Beijing\\
\email{xueyz19@mails.tsinghua.edu.cn, chenbinghui@bupt.cn, cangyu.gyf@alibaba-inc.com, xingtong.xxs@taobao.com, jschen@ustb.edu.cn, hbma@mail.tsinghua.edu.cn}}

\maketitle

\begin{abstract}
Customized generative text-to-image models have the ability to produce images that closely resemble a given subject.
However, in the context of generating advertising images for e-commerce scenarios, it is crucial that the generated subject's identity aligns perfectly with the product being advertised.
In order to address the need for strictly-ID-preserved advertising image generation, we have developed a Control-Net based customized image generation pipeline and have taken earring-model advertising as an example.
Our approach facilitates a seamless interaction between the earrings and the model's face, while ensuring that the identity of the earrings remains intact.
Furthermore, to achieve a diverse and controllable display, we have proposed a multi-branch cross-attention architecture, which allows for control over the scale, pose, and appearance of the model, going beyond the limitations of text prompts.
Our method manages to achieve fine-grained control of the generated model's face, resulting in controllable and captivating advertising effects.
\end{abstract}

\section{Introduction}
\label{sec:intro}

Generative text-to-image models have demonstrated remarkable capabilities in generating a wide range of high-quality images solely from user-defined text prompts~\cite{glide}~\cite{imagen}~\cite{sd}.
The ability to create novel images of subjects that resemble users' personalized items is truly captivating~\cite{dreambooth}~\cite{custom-diffusion}~\cite{anydoor}.
Nevertheless, in certain situations, achieving a mere \textit{similar appearance} falls short of fulfilling expectations.

In e-commerce scenarios, which play a crucial role in our daily lives and have a high demand for advertising image generation, sellers may opt to utilize generative models to produce advertising images for their products.
It is essential to maintain the identity of the product, ensuring that there are no noticeable alterations in its size, shape, structure, and color. Any significant changes may raise suspicions of false advertising.
This requirement for generation is referred to as \textit{strict-ID-preservation}.

\begin{figure*}[t]
    \centering
    \includegraphics[width=0.9\linewidth]{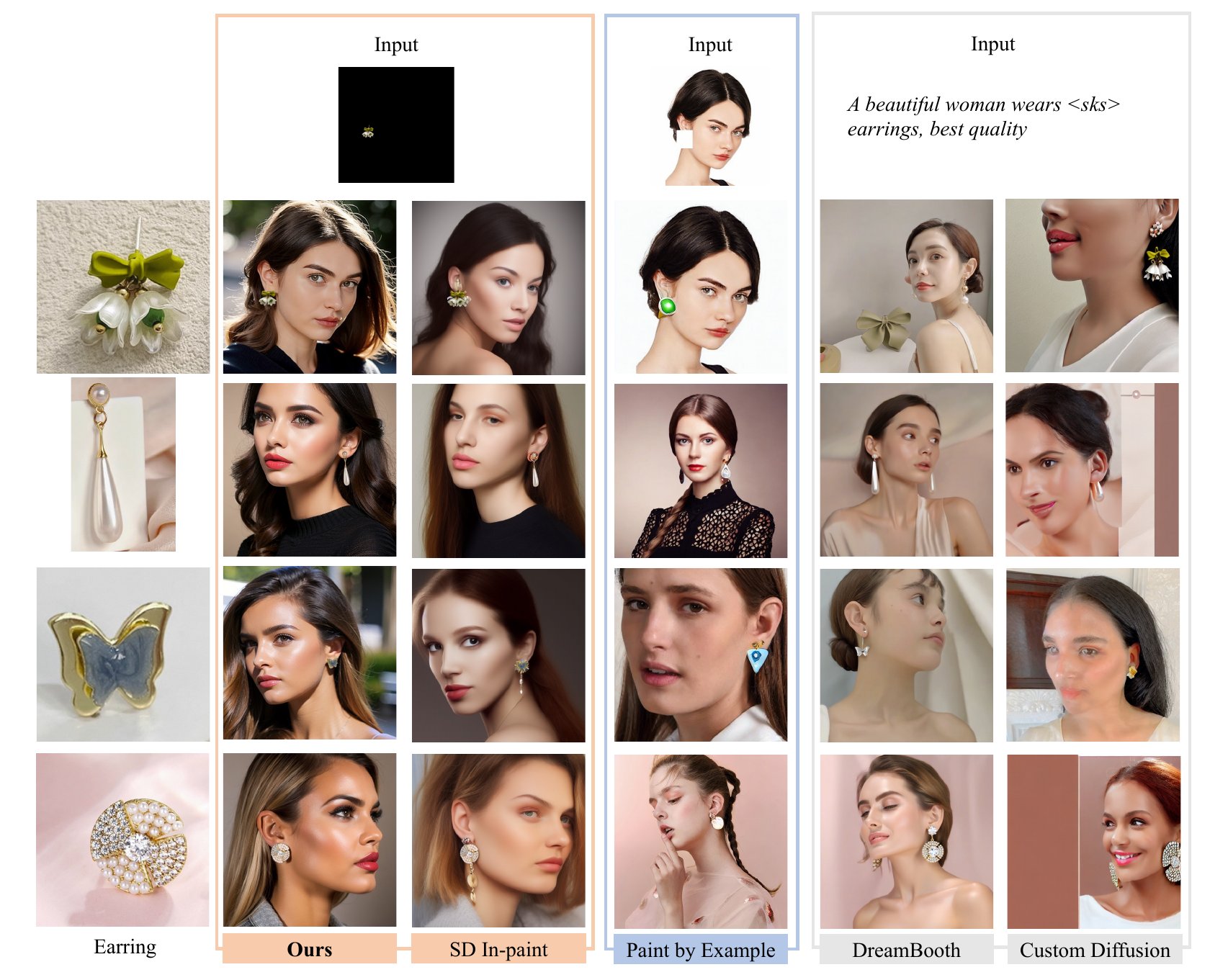}\vspace{-1em}
    \caption{The visualization comparison between our method with baselines. The first row presents examples of input for each method. Our method and Stable Diffusion (SD) in-painting~\cite{sd} use the earring image as the foreground and in-paint the background. Paint-by-Example~\cite{paint-by-example} in-paints the earring area on an earring-removed background.
    DreamBooth~\cite{dreambooth} and Custom Diffusion~\cite{custom-diffusion} are tuning-based text-to-image methods, where a text prompt with a special identifier $\langle sks \rangle$ is used as input.}
    \label{fig:main}\vspace{-1em}
\end{figure*}

\begin{figure}[t]
    \centering
    \includegraphics[width=0.65\linewidth]{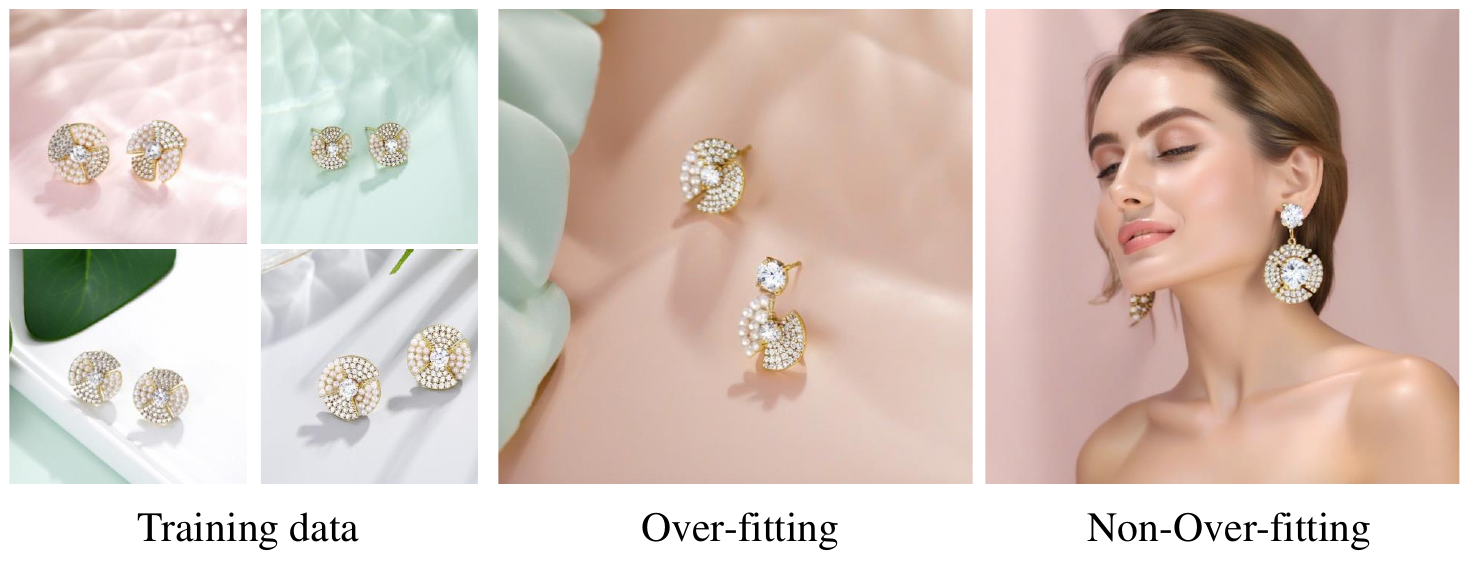}\vspace{-1em}
    \caption{The illustration of the over-fitting issue for the tuning-based customized generative models.}\vspace{-1em}
    \label{fig:overfitting}
\end{figure}

The existing customized generative models currently do not meet the requirement of \textit{strict-ID-preservation}.
\cref{fig:main} presents several examples of the earring-model advertisement generation.
These models are only capable of generating appearances that are \textit{similar but not identical} to the given reference.
While they may be suitable for personalized creation or daily use, their modifications on the appearance of commodities are unacceptable for advertising purposes.
The exemplar-based method~\cite{paint-by-example} decodes high-level image features during inference, resulting in a significant loss of information about the original commodity.
On the contrary, tuning-based methods~\cite{dreambooth}~\cite{custom-diffusion} require multiple reference images of the same subject and take several minutes to fine-tune in order to achieve satisfactory generation quality for each subject, which is inconvenient for users.
Additionally, the tuning-based methods suffer from the over-fitting issue, as demonstrated in \cref{fig:overfitting}, necessitating a labor-intensive search for appropriate hyper-parameters to resolve.
A more promising approach to achieve \textit{strict-ID-preservation} is to utilize in-painting, where the commodity's picture serves as the foreground and a suitable background is in-painted~\cite{sd}.
However, directly employing pretrained in-painting models may result in unreasonable and unnatural extensions of the foreground, thereby impacting the ID-preservation of the commodity.

To address the aforementioned issues, we present a novel pipeline for generating \textit{strictly-ID-preserved} advertising images.
Specifically, we focus on earrings, a popular commodity on e-commerce platforms, and believe that our approach is also enlightening to generating advertisements for other types of accessories.
Inspired by the spatial-conditioning ability of Control-Net~\cite{controlnet}, we utilize Control-Net as the underlying architecture and use the image of the earring as the conditioning image.
By training the pipeline on earring-model images, we teach it to generate a model face that seamlessly interacts with the earring placed at arbitrary positions.
We observe that Control-Net outperforms the naive in-painting approach in terms of spatial alignment, resulting in better preservation of the earring's shape.
Additionally, since the effectiveness of accessory advertising heavily relies on the expressiveness of the faces, users may desire diverse and fine-grained control over the generated models. 
Thus, we propose a multi-branch cross-attention architecture to replace the text-based cross-attention layers in Control-Net, which allows for fine-grained control over the \textbf{scale, pose, and appearance} of the model.
To ensure a balanced influence from different branches, we introduce a standard-deviation based normalization (STD-Norm) mechanism and a time-dependent weighting (TDW) strategy. 
These techniques enable our method to generate \textit{strictly-ID-preserved} earring-model advertising images while providing users with diverse and precise control over the generated models.
Contributions of this paper can be summarized as:
\begin{enumerate}
    \item We define a new task of \textit{strictly-ID-preserved} advertising image generation and propose a Control-Net based pipeline for it on earring images.
    \item We propose a multi-branch cross-attention architecture for the combination of the scale, the pose and the appearance control of the generated face.
    \item We propose branch balancing strategies STD-Norm and TDW for the multiple controls so that all branches work well for the generation results.
    
\end{enumerate}

\section{Related Works}
\label{sec:related_works}

\noindent \textbf{Diffusion Models:}
Deep generative models have achieved great success in generating high-quality images~\cite{biggan}~\cite{style}~\cite{dalle}~\cite{dalle2}~\cite{sd}.
Among these models, diffusion models~\cite{diff}~\cite{diff-beat-gan} are particularly noteworthy due to their recent advancements.
Diffusion models approach image generation as a multi-step denoising process, where an image contaminated with Gaussian noise is gradually cleaned.  
The training objective is to estimate the noise added onto the image.
In the Denoising Diffusion Probabilistic Model (DDPM)~\cite{ddpm}, the training objective was re-weighted and simplified, as shown in \cref{eq:ddpm}.
Here, $\mathbf{x}_0$ and $\mathbf{x}_t$ represent the clean and contaminated images respectively, $\epsilon$ denotes the noise and $t$ represents the time step. 

\begin{equation}
    L(\theta) = \mathbb{E}_{t,\mathbf{x}_0,\epsilon}[{||\epsilon - {\epsilon}_{\theta}(\mathbf{x}_t, t)||}^2]
    \label{eq:ddpm}
\end{equation}

In order to produce higher resolution images, researchers attempted to reduce computational complexity by employing multi-stage generation~\cite{cascaded} or exploring latent space~\cite{dalle2}.
The Latent Diffusion Model (LDM) ~\cite{sd}, which was subsequently expanded into Stable Diffusion (SD) and SDXL, enables the generation of images with dimensions of 512$\times$512 or even 1024$\times$1024.

\noindent \textbf{Conditional Diffusion Models:}
By incorporating textual control into diffusion models, conditional text-to-image models can generate images based on user-provided text prompts. 
In Stable Diffusion, text prompts are encoded into latent space using pretrained large language models like CLIP~\cite{clip} and then used as key-value pairs in cross-attention layers.
Additionally, other types of conditions such as layouts~\cite{layout}, poses~\cite{openpose}, edges~\cite{canny}, etc., are also explored.
Zhang and Agrawala introduced Control-Net, which injects additional conditions into pretrained large diffusion models~\cite{controlnet}. 
Due to the 2D formatting nature of the conditioning inputs, Control-Net conveniently injects images as conditions, ensuring strong spatial alignment between the condition and the generated image. 
We extend the cross-attention layers in Control-Net into multiple branches to enable diverse and fine-grained control over the generated models.
Recently, IP-Adapter adopted a similar text-image two-branch cross-attention architecture~\cite{ip-adapter}.
However, our method differs from IP-Adapter in two key aspects.
Firstly, we acknowledge that the model is more prone to over-fitting on the image branch when trained on small-scale task-specific training data.
Therefore, we propose a standard-deviation based normalization (STD-Norm) mechanism and a time-dependent weighting (TDW) strategy to balance the different branches.
Secondly, our method separates the pose and appearance of the model into distinct branches, allowing for more fine-grained control over the model for advertising purposes.

\noindent \textbf{Customized Generation:}
The concept of generating images of personalized subjects was first introduced in~\cite{dreambooth} and has since gained significant attention.
Typically, such customized generation methods involve using a set of reference images of a specific subject to fine-tune a pretrained large text-to-image model~\cite{dreambooth}~\cite{custom-diffusion}~\cite{cones}.
A unique text identifier is bound with the subject to indicate the model to generate images of that particular subject.
However, it should be noted that these methods, although capable of generating items with similar appearances to the reference, do not guarantee \textit{strict-ID-preservation} of the subject.
Additionally, fine-tuning the model and finding suitable hyper-parameters to avoid over-fitting often require a significant amount of time, ranging from several minutes to hours. 
Another approach to generating personalized content involves decoding high-level features of an exemplar during in-painting~\cite{paint-by-example}~\cite{objectstitch}.
However, this manner also fails to fully preserve the identity of the exemplar due to the lossy nature of encoding and decoding high-level features.

\begin{figure*}[t]
    \centering
    \includegraphics[width=0.95\linewidth]{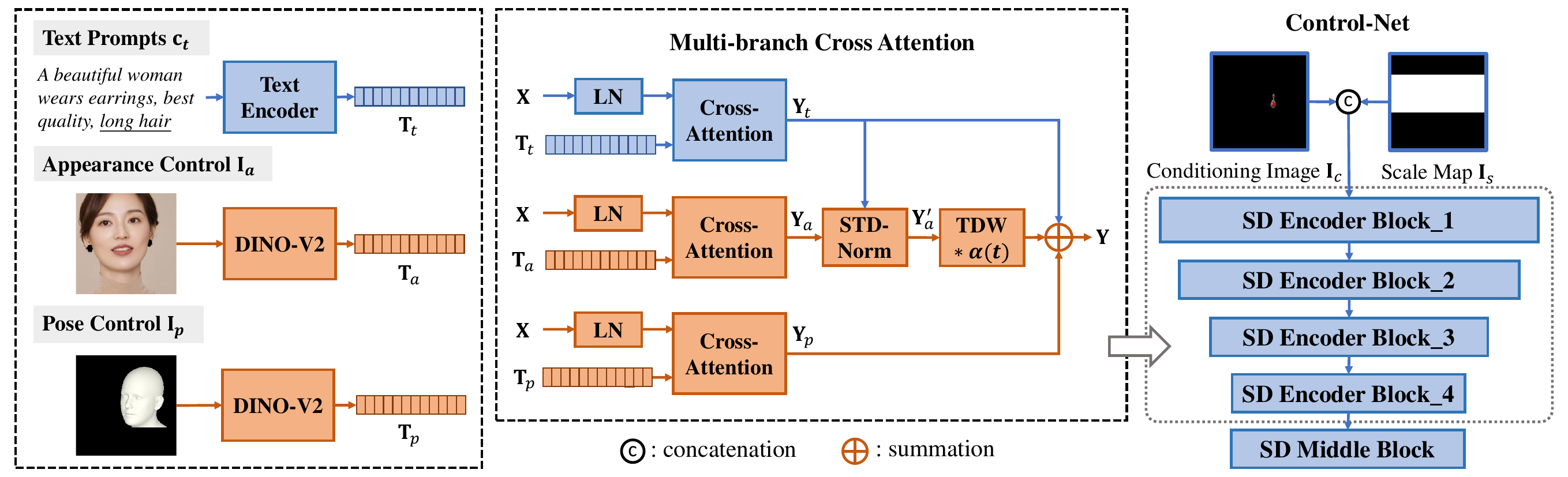}
    \caption{The overall pipeline of our method. By adding the standard-deviation based normalization (STD-Norm) and the time-dependent weighting (TDW), controls from different branches work well with each other.}
    \label{fig:pipeline}
\end{figure*}

\section{Methods}

The overall pipeline of our method is shown in \cref{fig:pipeline}.
In the subsequent part of this section, we provide detailed explanations of the key designs of the Control-Net based network.

\subsection{Strictly-ID-Preserved Generation}
For advertising purposes, we require that the item depicted in the generated image is identical to the user-provided reference.
As discussed in \cref{sec:intro}, existing methods for customized generation can only produce content that resembles the reference images, but the exact shape, structures, and colors may vary.
In contrast, in-painting methods offer more promise in preserving the identity of the product when fine-tuned on task-specific data.
Control-Net has demonstrated its ability to adapt pretrained diffusion models to task-specific conditions, even with a small task-specific dataset.
Additionally, the 2D formatting of the conditioning image ensures good spatial alignment between the condition and the generated result.
These characteristics make Control-Net highly suitable for achieving \textit{strict-ID-preservation}.

To train the Control-Net based pipeline, the earring is segmented out from the target image and used as the conditioning image $\mathbf{I}_c$.
The mask $\mathbf{M}$ indicates the corresponding earring area. 
We denote the conditioning image and other auxiliary control conditions introduced in the following sections as $\mathbf{c}$, and the text prompts as $\mathbf{c}_t$. Therefore, the training objective of the Control-Net is as follows:

\begin{equation}
    L(\theta) = \mathbb{E}_{t,\mathbf{x}_0, \mathbf{c}_t, \mathbf{c},\epsilon}[{||\epsilon - {\epsilon}_{\theta}(\mathbf{x}_t, t, \mathbf{c}_t, \mathbf{c})||}^2]
    \label{eq: control}
\end{equation}

During inference, we replace the area of the generated image where the earring is located with the reference earring image to insure \textit{strict-ID-preservation}, as depicted in \cref{eq: inference}.
$\mathbf{I}_g$ represents the generated image, while $\mathbf{I}_f$ represents the final result.  

\begin{equation}
    \mathbf{I}_{f} = \mathbf{I}_c + \mathbf{I}_g \odot (1-\mathbf{M})
    \label{eq: inference}
\end{equation}

\subsection{Scale Control}
To accurately represent the earring's size in the real world, the users need the option to adjust the relative scale between the model face and the earring in the generated image.
The pixel-scale of the earring is determined by the conditioning image, so the challenge lies in controlling the scale of the generated model.
One approach is to use text prompts such as `small face' or `big face'. 
However, describing the precise scale of the model face using natural language is challenging.
Considering that Control-Net effectively aligns the condition and the generation in terms of spatial positioning, we propose incorporating the scale condition into the conditioning image.

During training, we utilize the 2D key points of the human face to determine the upper and lower limits of the face, and create a binary scale map $\mathbf{I}_s$ based on these limits.
The scale map is an image where only the areas within the upper and lower limits of the face are assigned a value of 1.
Then $\mathbf{I}_s$ is appended to the conditioning image as an additional channel.
By doing so, the height and vertical position of the model face are explicitly incorporated into the Control-Net pipeline, while allowing the network to independently determine the horizontal position and shape of the model face.
During inference, the scale map can be provided by the user or deduced from the earring's position.
Further details can be found in the supplementary materials.

\subsection{Multi-Branch Cross-Attention}

To achieve visually appealing advertising effects, we wish to have diverse and fine-grained control over the generated model.
We consider the pose and appearance of the face as two crucial aspects to control.
Regarding the representation of the pose, in \cite{controlnet} the open-pose~\cite{openpose} 2D key points and limbs are overlaid on the conditioning image to control the pose of the generated person.
However, this approach also determines the position of the subject based on the key points, which is not desirable for our applications. 
In our case, the position of the model face needs to depend on the position of the earring.
To address this issue, we propose using the rendering of a human mesh model to represent the pose~\cite{smpl-x}.
By adjusting the corresponding mesh model parameters, we can easily normalize the shape, scale, position, and facial expressions of the human mesh model, ensuring that the rendering only depends on the pose.
We refer to this template rendering as $\mathbf{I}_p$.
During training, we we utilize OSX~\cite{osx} to estimate the SMPL-X~\cite{smpl-x} mesh model parameters of the reference earring-model image.
The mesh model is then adjusted so that the nose is rendered at the center of the image, with the default depth.
Additionally, the shape and expression parameters are set to 0 to eliminate their effects.
During inference, $\mathbf{I}_p$ of any desired pose can be produced by adjusting the pose parameters of the mesh model.

As for the appearance, we use a model face image $\mathbf{I}_a$ as the representation.
During training, we use the earring-removed target image as $\mathbf{I}_a$.
Since $\mathbf{I}_a$ also contains information about the position, scale, and pose of the model, we apply random scaling, random rotation, and random horizontal flipping to it during training to separate the appearance information from these interfering factors.
Furthermore, we randomly drop the appearance branch during training to enable the pipeline to generate the appearance solely based on the text prompts.
During inference, any face image can be chosen as $\mathbf{I}_a$, or the appearance branch can be omitted.

Since $\mathbf{I}_p$ and $\mathbf{I}_a$ are not spatially aligned with the earring-model image, it is not appropriate to inject them into Control-Net via the conditioning image.
Inspired by the injection of textual control, we propose a multi-branch cross-attention architecture.
For each cross-attention layer in the original Control-Net encoder blocks, we implement two additional structurally identical cross-attention layers for pose and appearance control respectively.
Referring to the structure of the textual cross-attention layer, the text prompts $\mathbf{c}_t$ are first embedded into latent space to obtain the embedded control tokens $\mathbf{T}_t$.
Then, Layer-Norm(LN) and Cross-Attention(CA) are applied to the output $\mathbf{X}$ from the previous layer and the embedded control tokens to generate the residual $\mathbf{Y}_t$, as is depicted in \cref{eq:ca-1}. 


\begin{equation}
    \mathbf{Y}_t = CA(LN(\mathbf{X}), \mathbf{T}_t)
    \label{eq:ca-1}
\end{equation}


In the CA operation, the normalized input is projected into query tokens $\mathbf{Q}$, and the embedded control tokens are used to generate the key tokens $\mathbf{K}$ and value tokens $\mathbf{V}$.
A linear projection layer is employed to produce the output.

\begin{equation}
    CA(LN(\mathbf{X}), \mathbf{T}_t) = Linear(Softmax(\frac{\mathbf{Q}\mathbf{K}^T}{\sqrt{d}})\mathbf{V})
    \label{eq:CA-1}
\end{equation}

\begin{equation}
    \mathbf{Q}=LN(\mathbf{X})\mathbf{W}_q, \mathbf{K}=\mathbf{T}_t \mathbf{W}_k, \mathbf{V}=\mathbf{T}_t \mathbf{W}_v
    \label{eq:CA-2}
\end{equation}

The pose and appearance branches operate in the same manner, with the only difference being the manner to acquire the embedded control tokens.
We utilize  DINO-V2~\cite{dinov2}, a powerful visual feature extractor, to extract the features of $\mathbf{I}_p$ and $\mathbf{I}_a$.
The final classification token and the patch tokens in the DINO-V2 output are concatenated, resulting in 257 tokens for the pose and appearance control respectively.
Then additional linear projections are employed to project the visual tokens to the same dimension as $\mathbf{T}_t$.
Following the multi-branch cross-attention layers' processing, three residuals, namely $\mathbf{Y}_t$, $\mathbf{Y}_p$, and $\mathbf{Y}_a$, are obtained.

\begin{equation}
    \mathbf{T}_p = DINO(\mathbf{I}_p)\mathbf{W}_p, \mathbf{T}_a = DINO(\mathbf{I}_a)\mathbf{W}_a
\end{equation}


\subsection{Branch Balancing}
\label{subsec:branch}
In a previous work, IP-Adapter adopts a similar text-image two-branch cross-attention architecture to combine textual and visual control~\cite{ip-adapter}.
The outputs of the two branches are combined using weighted summation, where the weights can be manually adjusted to strike a balance between visual alignment and textual alignment.
When trained on 10 million text-image pairs covering various topics~\cite{laion}~\cite{coyo}, this combination performs well in inheriting both controls. 
However, in our specific scenario, where the training data for the task is limited ($\sim$0.23 million) and the data context is relatively monotonous, severe over-fitting to the appearance branch occurs. 
This makes it challenging to manually determine appropriate weights.
We have observed that during over-fitting, the standard deviation of the appearance branch significantly exceeds that of the other branches ($\sim$5$\times$). 
As a result, the effects of the other branches become obscured.
To address this issue and adaptively balance the effects of the appearance branch with the others, we propose a mechanism called standard-deviation based normalization (STD-Norm), which is depicted in \cref{eq:norm}.
It involves calculating the standard-deviation of the textual branch output and then normalizing and adjusting the appearance branch output to have the same standard-deviation as $\mathbf{Y}_t$.

\begin{equation}
    \mathbf{Y'}_a = \frac{\mathbf{Y}_a}{std(\mathbf{Y}_a)} * std(\mathbf{Y}_t)
    \label{eq:norm}
\end{equation}

Moreover, it has been observed that even when the STD-Norm mechanism is applied, the generated image still occasionally fails to accurately generate text-controlled elements such as glasses and hats.
Inspired by the coarse-to-fine nature of the denoising process, we suggest that the early stage of denoising should prioritize generating global structures through textual control, while appearance details should be injected later in the process. 
To achieve this, we introduce a time-dependent weighting (TDW) factor to the appearance branch and the combination of multiple branches finally becomes as depicted in \cref{eq:re-weight}.
Empirically, a truncated linear changing strategy controlled by a hyper-parameter $\gamma$ works well, as is shown in \cref{eq:alpha}.

\begin{equation}
    \mathbf{Y} = \mathbf{Y}_t + \mathbf{Y}_p + \alpha(t) \cdot \mathbf{Y'}_a
    \label{eq:re-weight}
\end{equation}

\begin{equation}
    \alpha(t) = 
    \begin{cases}
        0,& {t\geq2\gamma}\\
        1 - \frac{t-\gamma}{\gamma}, & {\gamma<t<2\gamma} \\
        1,& {t\leq\gamma}
    \end{cases}
    \label{eq:alpha}
\end{equation}  

\section{Experiments}

In this section, we demonstrate our method's ability of generating \textit{strictly-ID-preserved} and controllable earring-model advertising images.

\begin{table*}[]
\centering
\caption{The method comparison between ours and the baselines. PbE, DB and CD are short for Paint-by-Example, DreamBooth and Custom Diffusion respectively.}
\begin{tabular}{l|ccccc}
\hline
                                & SD Inpaint~\cite{sd} & PbE~\cite{paint-by-example} & DB~\cite{dreambooth} & CD~\cite{custom-diffusion} & Ours \\ \hline
Single Reference                & \checkmark         & \checkmark              & $\times$         & $\times$               & \checkmark  \\
Strict-ID-Preservation          & $\times$          & $\times$               & $\times$         & $\times$               & \checkmark  \\
Textual Control                 & \checkmark         & $\times$               & \checkmark        & \checkmark              & \checkmark  \\
Scale\&Pose\&Appearance Control & $\times$          & $\times$               & $\times$         & $\times$               & \checkmark  \\ \hline
\end{tabular}
\label{tab:method}
\end{table*}

\begin{figure*}[t]
    \centering\vspace{-1em}
    \includegraphics[width=0.95\linewidth]{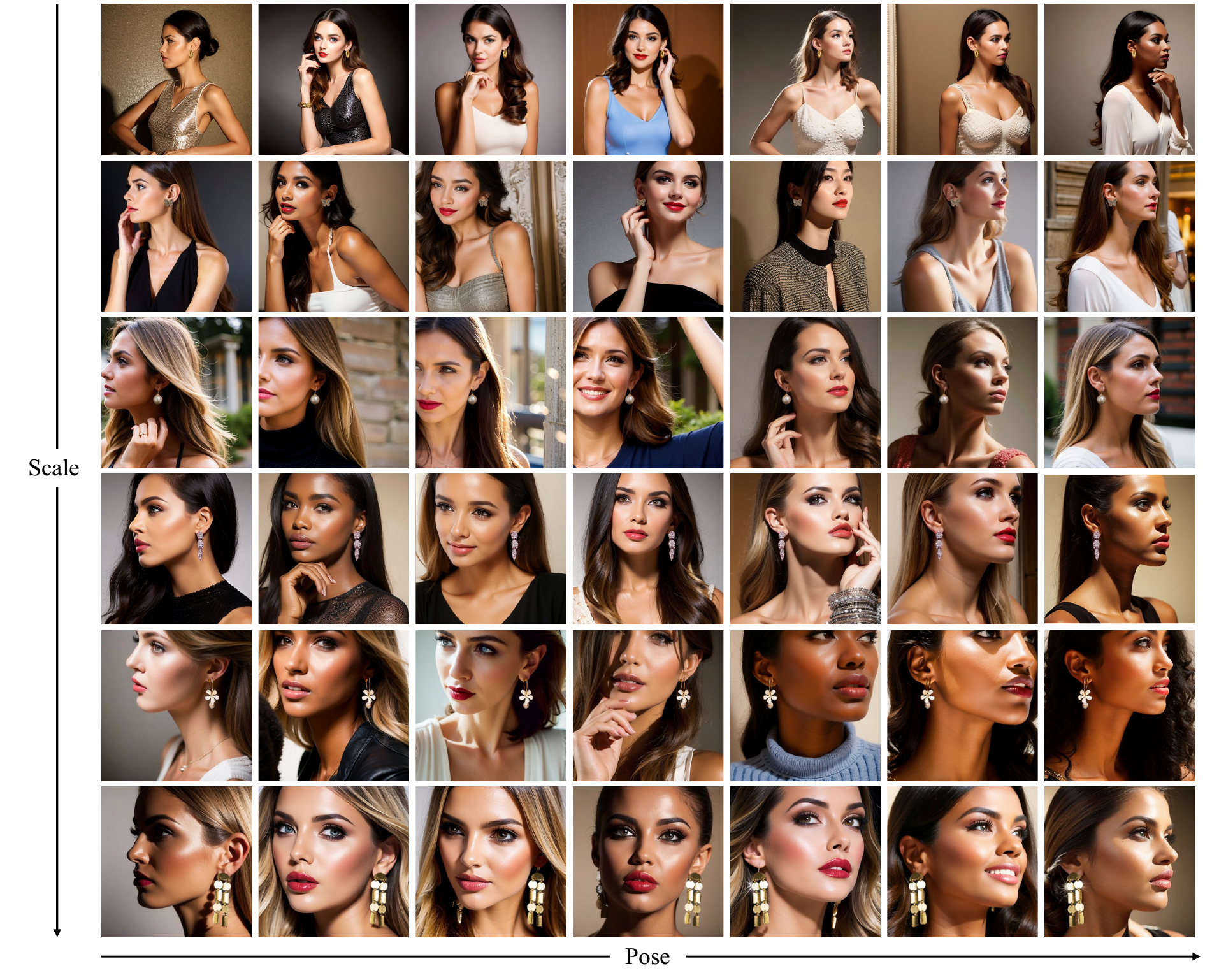}\vspace{-1em}
    \caption{The illustration of the scale and pose control. For each column of the figure, the pose control is fixed and the scale control changes from small to large. And for each row of the figure, the scale control and the earring image are fixed, whereas the pose control changes from facing left to facing right.}
    \label{fig:scale&pose}\vspace{-1em}
\end{figure*}


\begin{figure*}[t]
    \centering\vspace{-1em}
    \includegraphics[width=0.95\linewidth]{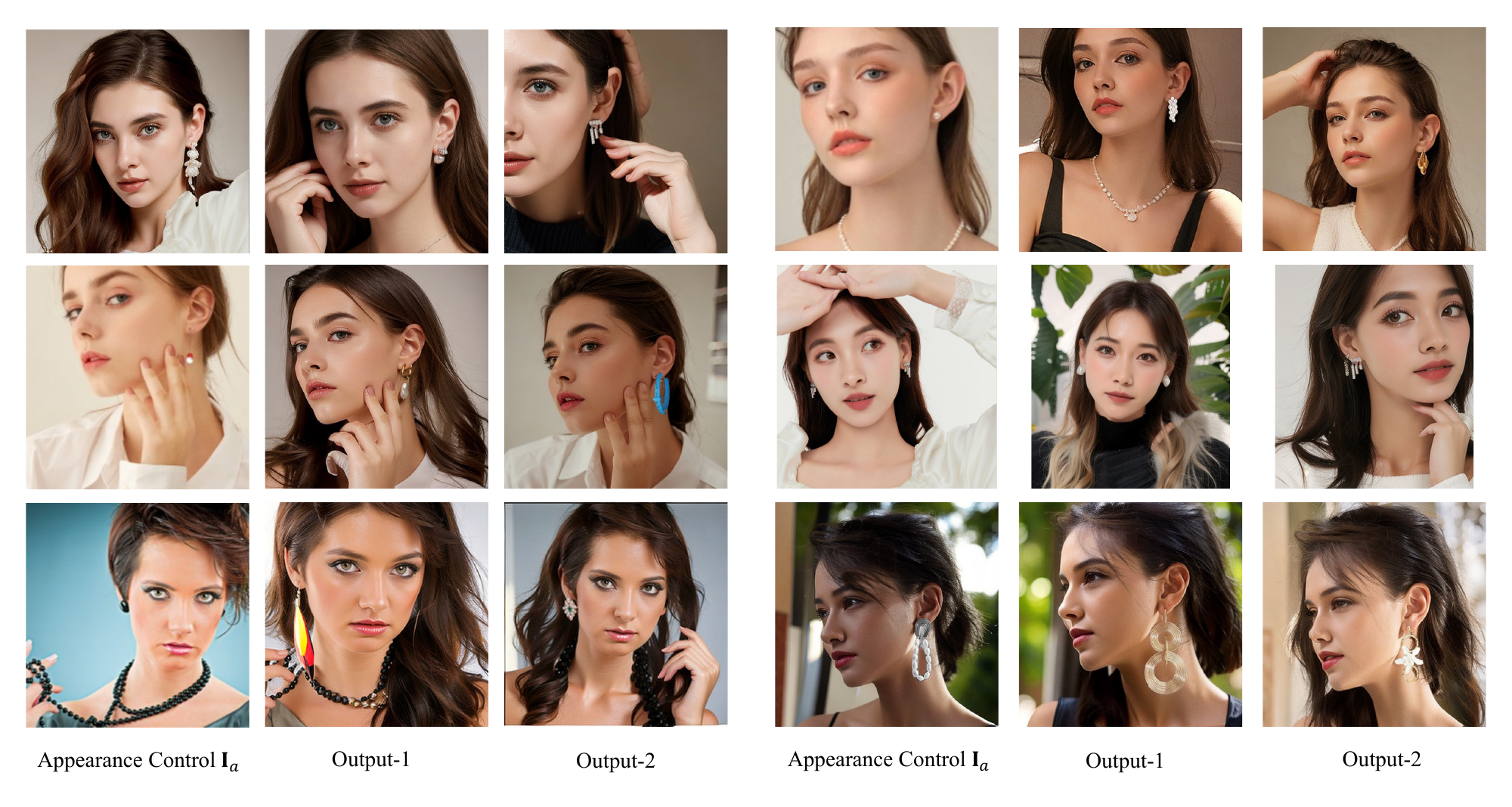}\vspace{-1em}
    \caption{The illustration of the appearance control. For each appearance control $\mathbf{I}_a$, we select two different earrings and generate two images based on the earring and $\mathbf{I}_a$.}
    \label{fig:appearance}\vspace{-1em}
\end{figure*}

\begin{figure*}[t]
    \centering\vspace{-1em}
    \includegraphics[width=0.95\linewidth]{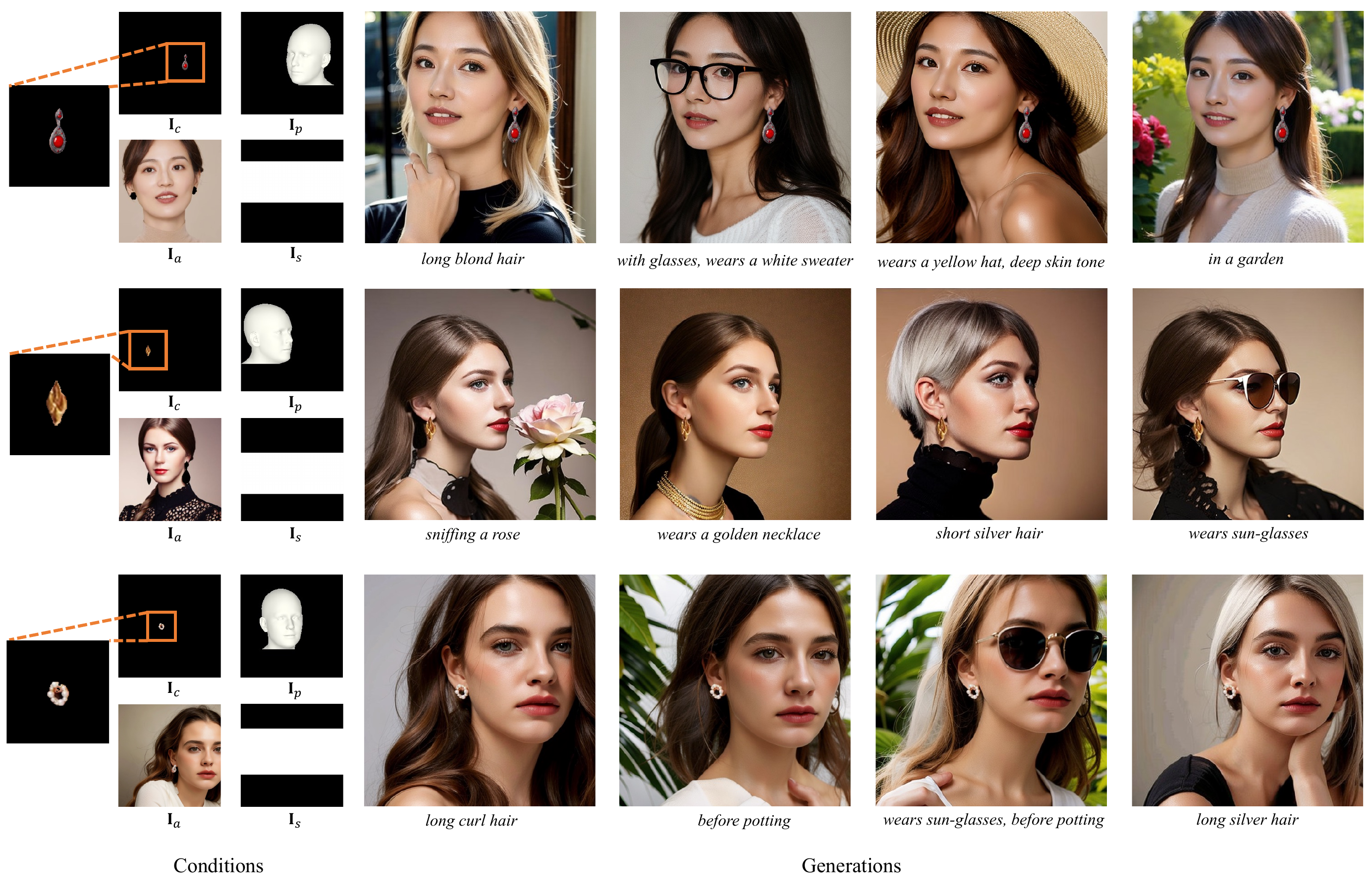}\vspace{-1em}
    \caption{Combination of the scale, pose, appearance and textual control. The text in slanting typeface is the additional text prompt. Our method shows very good results both in terms of the ID-preservation and the combinatorial effects of various controls.}
    \label{fig:text&appearance}\vspace{-2em}
\end{figure*}

\noindent \textbf{Dataset:} To build the training dataset, we collected a total of 230K earring-model images from the internet.
Each image was then processed using BLIP~\cite{blip} to extract a caption. We used Segment-Anything~\cite{segment-anything} to segment the earrings from the images, and OSX~\cite{osx} to estimate the SMPL-X parameters of the model and the corresponding camera parameters for each image.
Once we obtained the mesh parameters, we computed the 3D coordinates of 68 key points on the human face and projected them onto the image plane to obtain the 2D key points. 
For evaluation purposes, we collected an additional 3000 earring-model images and pre-processed the data following the same procedure as the training data.
During inference, unless otherwise specified, a default caption of \textit{`A beautiful woman wears earrings, best quality'} was used.

\noindent \textbf{Implementation details:} We adopted Stable Diffusion V1.5 (SD) as the backbone architecture and initialized it with the pretrained weights provided by \textit{majicMIX}~\cite{majicMIX}. 
The conditions $\mathbf{I}_p$ and $\mathbf{I}_a$ were resized to $224 \times 224$ before being fed into DINO-V2 to extract features.
The Control-Net was trained on 8 NVIDIA V100 GPUs for 20 epochs, with a learning rate of 1e-5. 
During inference, we used the DDIM sampler~\cite{ddim} for 30 steps and set the classifier-free guidance~\cite{cfg} to 7.0.
The hyper-parameter $\gamma$ was chosen as 400 for the time-dependent weighting factor.

\noindent \textbf{Baselines}: 
Depending on the number of reference images required, the baselines can be categorized into two groups.
The first includes SD in-painting and Paint-by-Example~\cite{paint-by-example}, which only require a single reference image.
The second group includes tuning-based methods like DreamBooth~\cite{dreambooth} and Custom Diffusion~\cite{custom-diffusion}, which require a minimum of 3 reference images to fine-tune the model.
\cref{tab:method} illustrates the comparison between our method and the baselines. 
It is worth noting that our method is the only one that can achieve \textit{strict-ID-preservation} and offer diverse means of control over the generated model.

\subsection{Visualization of the Generation}

\cref{fig:scale&pose}, \cref{fig:appearance} and \cref{fig:text&appearance} present the visualization results of our method.
It is clear to note that the generated images preserve the shape, the size and the appearance of the reference earrings very well.
Additionally, the interaction between the earring and the model is highly faithful.
As for the control effects, we demonstrate the scale and pose control separately.
In \cref{fig:scale&pose}, the input pose control changes from left to right for each row of the figure and the input scale control changes from small to large for each column.
The appearance branch weighting factor $\alpha(t)$ is set to zero in the aforementioned experiments, allowing for the free generation of the model face.
It is worth noting that the scale and pose of the generated model faces align remarkably well with the specified conditions, showcasing the diverse and fine-grained control capability of our method.
In \cref{fig:appearance}, we show several examples of the appearance control images and the generated outputs based on them.
It is clear to see that the appearance images control the face appearance of the generated images very well. 
Quantitative analysis on the human face identity will be presented in \cref{subsec:ablation}.

When combined with appearance control and additional textual prompts, our method can generate even more diverse and captivating advertising effects. 
\cref{fig:text&appearance} presents several examples of the combination of multiple controls.
The results show that the scale, the pose, the appearance and the text-based details are decoupled as desired.
Therefore, the users can easily make any desired combination of these factors to facilitate advertising of their products.
Noted that the appearance control images are not included in the training data.
This indicates that our method is able to generalize to unseen faces as well. 

\subsection{Comparisons with Baselines}


\noindent \textbf{Single Reference:}
To compare our method with baselines that rely on a single reference, we conducted quantitative experiments on the test data.
For each of the 3000 test images, we segmented out the earring as the conditioning image. 
The scale map $\mathbf{I}_s$, pose control $\mathbf{I}_p$, and appearance control $\mathbf{I}_a$ were prepared based on the earring-removed background.
For the SD in-painting method, we used the reference earring image as the foreground and applied the in-painting mode of SD V1.5 to fill in the background.
As for Paint-by-Example, the reference earring image was used as the exemplar and the model in-painted the area within the bounding box of the earring in the original image based on it.


\begin{table}[]
\centering
\caption{The quantitative comparisons between our method with SD in-painting and Paint-by-Example.}
\begin{tabular}{l|ccc}
\hline
                 & FID $\downarrow$   & Mask IoU $\uparrow$ & CLIP-S $\uparrow$ \\ \hline
SD In-paint      & 17.92 & 36.5     & 83.6       \\
Paint-by-Example & \textbf{4.42}  & 33.4     & 70.9       \\
Ours             & 4.44  & \textbf{87.3}     & \textbf{95.0}       \\ \hline
\end{tabular}
\label{tab:single}
\end{table}

To measure the overall quality of the generated images, we utilize the FID score~\cite{fid} as a metric, comparing the generated images with the test images.
Additionally, to assess the preservation of the earring's identity, we randomly select 300 images from the test dataset and manually annotate the segmentation mask of the earring in the generated images using Segment-Anything.
Then the Intersection-over-Union (IoU) of the masks between the generation and the ground truth is calculated to measure the retention of the earring's shape.
Moreover, the mask is employed to determine the bounding box of the earring, and the CLIP score (CLIP-S)~\cite{clip} of the bounded area is computed as an indicator of appearance preservation.
\cref{tab:single} presents the quantitative results and \cref{fig:main} shows several visualization examples.
It is important to note that Paint-by-Example utilizes the earring-removed real image as the background, hence a high FID score is expected.
Nevertheless, our method achieves a comparable FID score, demonstrating the high fidelity of the images generated by our approach.
Regarding ID-preservation, our method significantly outperforms the baselines in terms of mask IoU and CLIP-S.

\noindent \textbf{Multiple References:}
Tuning-based customized generative models typically require a set of reference images for fine-tuning.
To compare with them, we collected 5 sets of images, each containing 5 images of the same earring.
Then DreamBooth and Custom Diffusion were applied to each set of images.
We utilized the codes from $\textit{diffusers}$~\cite{diffusers} to train the model and discovered empirically that it is difficult to prevent the tuned model from over-fitting to the training data.
We define over-fitting as the generation of only the earring without the model face as is shown in \cref{fig:overfitting}. 
This may be due to the fact that the earring area is typically very small in an earring-model image, making it difficult for the model to associate the small earring with the model face, which occupies the majority of the image.
We employed the tuned models to generate 50 images for each earring and calculated the non-over-fitting rate (NOR).
For the non-over-fitting cases, we computed the CLIP-S using the same procedure aforementioned.
As for our method, we imitated the users' practice of randomly placing the earring image on a black background as the conditioning image, and randomly determining the scale and pose control.
Details can be referred to in the supplementary materials.
\cref{tab:multi} shows the quantitative comparisons and the visualization results are shown in \cref{fig:main}, clearly demonstrating that the tuning-based methods fail to achieve \textit{strict-ID-preservation}, whereas our method successfully produces \textit{strictly-ID-preserved} earring-model images with high fidelity.

\begin{table}[t]
\centering
\caption{The quantitative comparisons between our method with DreamBooth and Custom Diffusion.}
\vspace{-1em}
\begin{tabular}{l|cc}
\hline
                 & CLIP-S $\uparrow$ & NOR $\uparrow$ \\ \hline
DreamBooth       & 73.7       & 32\%                 \\
Custom Diffusion & 75.6       & 42\%                 \\
Ours             & \textbf{94.6}          & -                    \\ \hline
\end{tabular}
\label{tab:multi}
\end{table}

\begin{table}[]
\centering
\caption{The control effectiveness comparisons between our method with textual control.}
\vspace{-1em}
\begin{tabular}{l|ccc}
\hline
                & Scale & Pose & Race \\ \hline
Textual Control & 73.7\%     & 60.7\%    & 41.0\%    \\
Ours            & \textbf{92.0\%}     & \textbf{96.3\%}    & \textbf{94.3\%}    \\ \hline
\end{tabular}
\label{tab:text}
\end{table}

\begin{figure}[t]
    \centering\vspace{-1em}
    \includegraphics[width=0.9\linewidth]{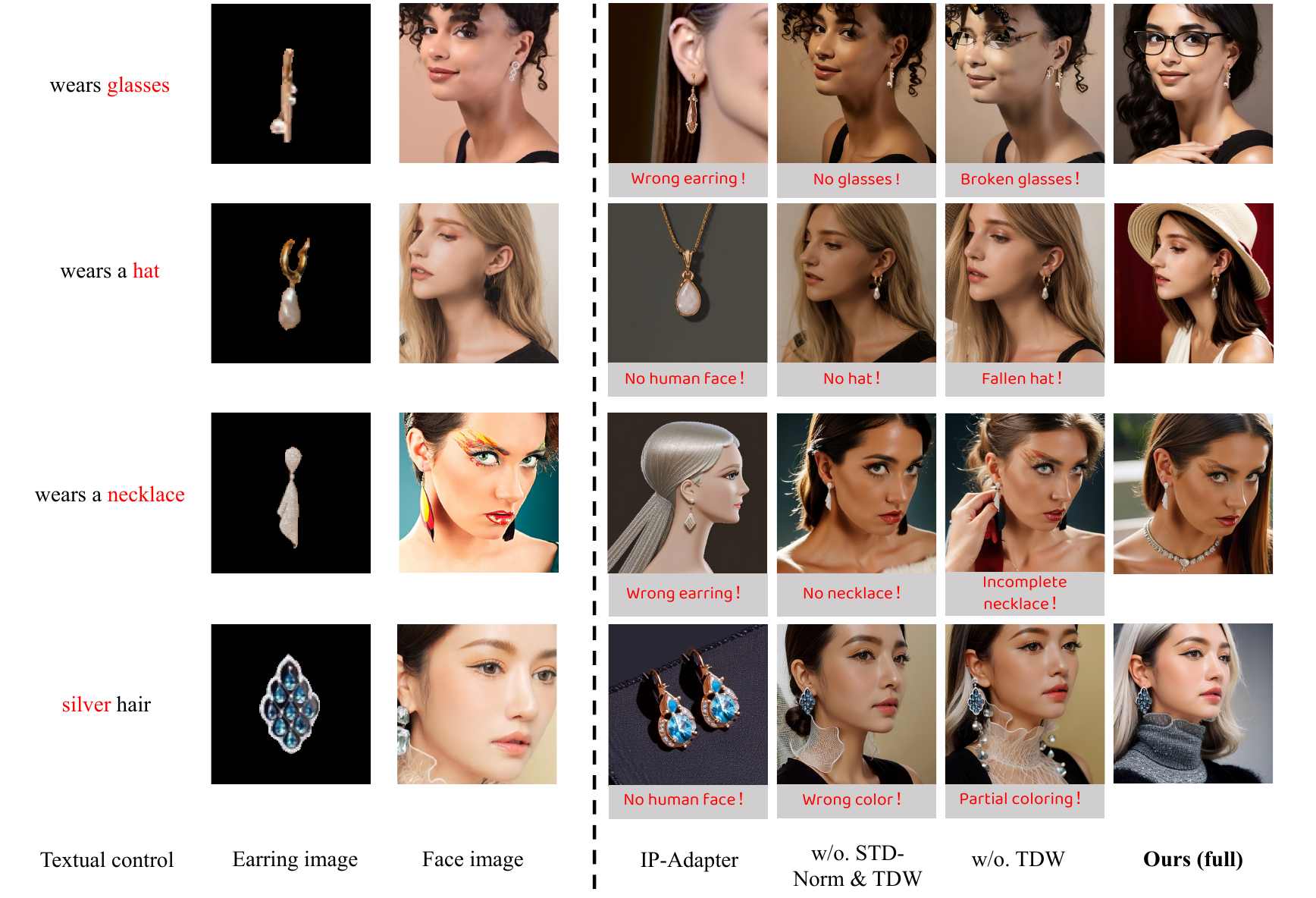}\vspace{-1em}
    \caption{The illustration of the effectiveness of branch balancing.}
    \label{fig:ablation}\vspace{-2em}
\end{figure}

\subsection{Analysis}
\label{subsec:ablation}
\noindent \textbf{ID-Preservation of the Human Face:}
To give a quantitative measurement of the human face's ID consistency, we generate 1000 images based on 50 subjects' faces and evaluate the identity of the generated human faces using the \textit{deepface}~\cite{deepface} tool.
98.3\% of them are verified as the same identity with the control, showing that our method preserves the human identity pretty well.
Moreover, we conduct a user study on the ID-preservation of the human face on these images.
20 volunteers are recruited to judge if the human identity in the generated image is the same as the appearance control image.
In total, 96.4\% of the cases are considered as the same identity, 3.1\% of the cases are considered as `uncertain', whereas only 0.5\% of the cases are judged as different identity.
It is good enough for our application since we only want to roughly control the faces' appearance rather than demand the generated face to be exactly the same with the control. 

\noindent \textbf{Comparisons with Textual Control:}
\label{subsec:text}
Our method offers a fine-grained and flexible approach to controlling the scale, pose, and appearance of a face.
To evaluate the effectiveness of our method compared to textual control, we annotate the face scale, face pose, and race of the model in the training data and train a text-based version for each factor.
The face scale and pose are determined using the mesh model, while the race is annotated using the \textit{deepface}~\cite{deepface} tool.
In summary, we define 3 classes of scale, 3 classes of pose, and 6 classes of race, and design textual prompts for each class.
For each factor, we randomly generate 300 images for our method and the text-based counterpart, and measure the alignment rate between the generated images' class label and the specified condition.
Details of the experimental implementation can be referred to in the supplementary materials.
The alignment rates are presented in \cref{tab:text}, which clearly demonstrate that our method enables more precise control over the generation compared to textual prompts.


\noindent \textbf{Design of Branch Balancing:}
In \cref{subsec:branch}, we introduce the STD-Norm scheme and the TDW strategy to balance the influence of the appearance branch and the other branches.
\cref{fig:ablation} presents several examples of using different choices of branch balancing strategy.
Noted that we also compare with IP-Adapter \cite{ip-adapter}.
However, directly using it leads to failures, e.g. failure to preserve the earring's identity or failure to generate a human face. It is because IP-adapter uses a image encoder to compress the reference object to high-level compact embedding and then generate images conditioned on this embedding. During this process, information loss will easily happen such that the object's ID cannot be strictly preserved, and no human features are introduced to ensure the generation of human. In contrast,
it is worth noting that both STD-Norm and TDW are essential for achieving effective control of all the branches.
Without both of them, the appearance branch will dominate the generated result, leading to a failure of textual control.
On the other hand, with only STD-Norm, the generation still occasionally suffers from broken contents prompted by the text. All these phenomenon demonstrate the importance of our STD-Norm and TDW strateties.
\vspace{-1em}
\section{Conclusion}
\vspace{-1em}
In this study, we present a Control-Net based pipeline for generating accessory advertising images, which ensures \textit{strict-ID-preservation} of the product for advertising uses.
Additionally, we introduce a multi-branch cross-attention architecture to provide fine-grained control over the scale, pose, and appearance of the generated model face, resulting in diverse and captivating advertising effects.
To maintain a balance between appearance control and other factors, a standard-deviation based normalization (STD-Norm) mechanism and a time-dependent weighting (TDW) strategy are proposed.
Extensive quantitative and qualitative experiments are conducted for earring-model image generation to demonstrate the superiority of our method.
Currently, our method achieves \textit{strictly-ID-preservation} by copying-and-pasting the earring image in the generated image, which leads to limitations of changing the rotation and the lighting conditions of the earring automatically.
This will be our future research emphasis.


%
%
\bibliographystyle{splncs04}
\bibliography{main}

\clearpage
\setcounter{page}{1}

\section{Supplementary}
\label{sec:rationale}

\subsection{Pipeline}

\begin{figure*}[thb]
    \centering
    \includegraphics[width=0.99\linewidth]{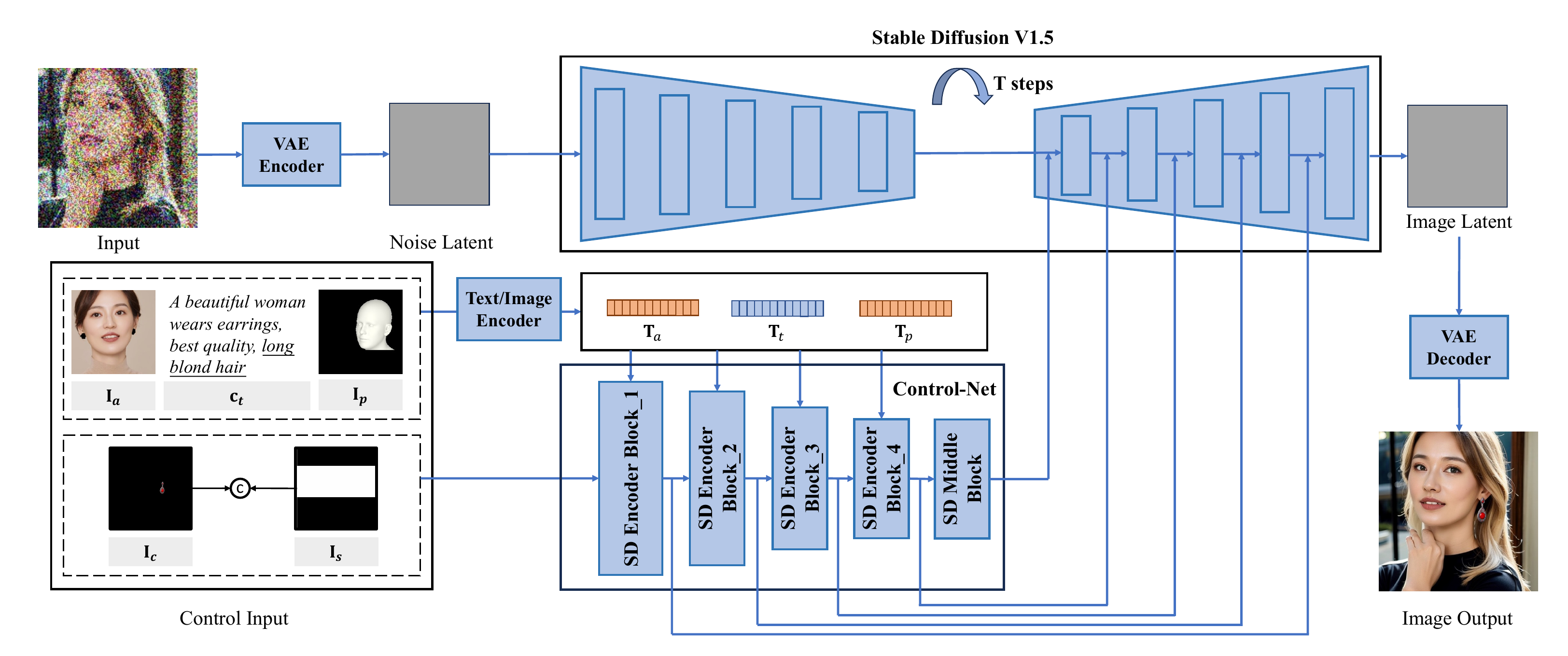}
    \caption{More detailed illustration of the whole pipeline.}
    \label{fig:pipeline_supp}
\end{figure*}

In \cref{fig:pipeline}, we provide an illustration of the pipeline and highlight the multi-branch cross-attention architecture.
Here we provide a more detailed illustration of the processing of the whole pipeline in \cref{fig:pipeline_supp}.
The pipeline uses SD V1.5 to generate the images and uses Control-Net to inject multiple controls.
The earring image $\mathbf{I}_c$ and the scale control $\mathbf{I}_s$ are concatenated as the conditioning image of the Control-Net.
And the appearance image $\mathbf{I}_a$, the text prompts $\mathbf{c}_t$ and the pose control $\mathbf{I}_p$ are first encoded into latent features $\mathbf{T}_a$, $\mathbf{T}_t$ and $\mathbf{T}_p$ using their corresponding feature encoders and then injected into the Control-Net via multi-branch cross-attention.
The structure of multi-branch cross-attention is illustrated in \cref{fig:pipeline}.

\subsection{Construction of $\textbf{I}_s$, $\textbf{I}_p$ and $\textbf{I}_a$}
\label{subsec:construct}
\textbf{Training:}

\begin{figure}[t]
    \centering
    \includegraphics[width=1.00\linewidth]{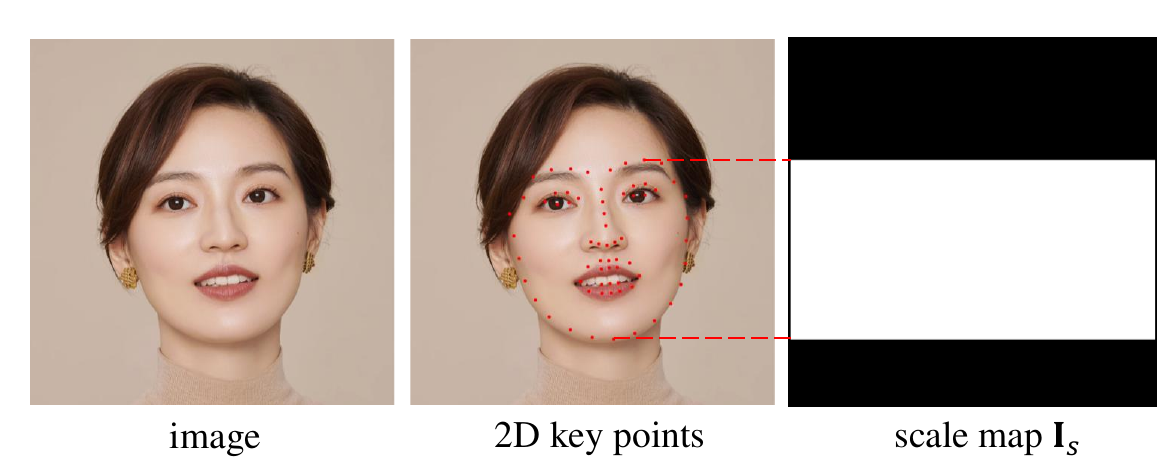}
    \caption{The illustration of the 2D face key points and the corresponding scale map for training.}
    \label{fig:supp_train}
\end{figure}

The scale map $\mathbf{I}_s$, pose control $\mathbf{I}_p$, and appearance control $\mathbf{I}_a$ are all prepared based on the target image for training purposes.
The target image is sent to OSX~\cite{osx} in order to estimate the SMPL-X~\cite{smpl-x} mesh parameters and camera parameters.
Using the mesh, the 3D coordinates of 68 human face key points can be calculated, and the 2D coordinates can be obtained by projecting these 3D key points onto the image plane using the camera parameters.
\cref{fig:supp_train} illustrates an example of the projected 2D face key points.
Based on these key points, the scale map $\mathbf{I}_s$ can be constructed as a binary image, where only pixels within the upper and lower limits of the key points are set to 1.

Regarding the pose control, the estimated SMPL-X mesh is first \textit{`normalized'} and then rendered onto a black image plane, referred to as $\mathbf{I}_p$.
The mesh normalization process consists of two steps.
Firstly, all parameters unrelated to the pose, such as shape parameters and facial expression parameters, are set to 0.
Secondly, the mesh is moved to a default coordinate, ensuring that the position and scale of the rendered mesh are independent of the data.
Empirically, moving the nose key point to (0, 0, 8) along with a camera using $\mathit{K}$ as the intrinsics has proven to be effective.

\begin{equation}
    K = \left [ \begin{array}{ccc}
        5000. & 0. & 128.\\
        0. & 5000. & 128.\\
        0. & 0. & 1.
        \end{array} 
        \right ]
    \label{eq: K}
\end{equation}

Regarding appearance control, we apply augmentation on the earring-removed image to create $\mathbf{I}_a$. 
This augmentation process involves three steps:

\begin{itemize}
    \item random rotation within $-30^\circ\sim30^\circ$.
    \item random horizontal flipping with a probability of 0.5.
    \item random resized crop with the cropped area ratio ranging from 0.25 to 1, while preserving the aspect ratio.
\end{itemize}

\noindent \textbf{Inference:}

\begin{figure}[t]
    \centering
    \includegraphics[width=1.00\linewidth]{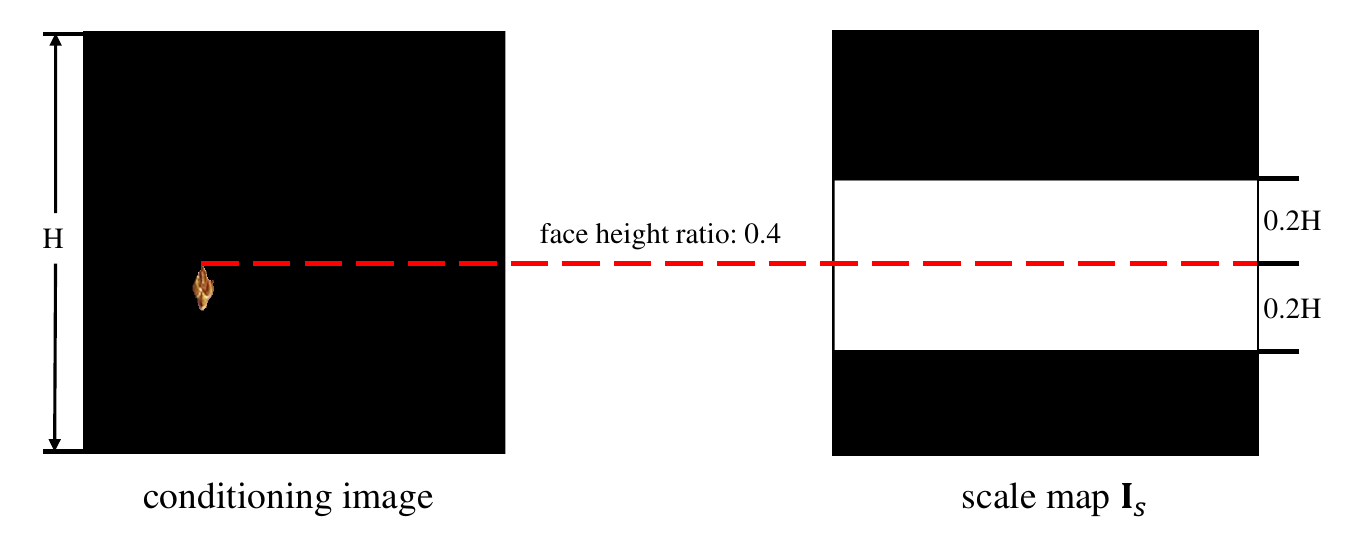}
    \caption{The illustration of the construction of the scale map for inference.}
    \label{fig:supp_inference}
\end{figure}

During the inference process, the scale map $\mathbf{I}_s$, the pose control $\mathbf{I}_p$, and the appearance control $\mathbf{I}_a$ are prepared based on the user's preferences.
For $\mathbf{I}_s$, the users have two options. They can either provide the actual size of the earring in the real world or directly determine the desired pixel-space height ratio between the face and the image.
In the first case, we calculate the pixel-space height of the face based on the pixel-space height of the earring, and the actual height ratio between the face and the earring.
While in the second case, the pixel-space height of the face is directly calculated based on the user-provided height ratio.
Once the pixel-space face height is determined, we construct $\mathbf{I}_s$ by placing the centering of the face area at the upper bound of the earring.
The process is illustrated in \cref{fig:supp_inference}.

\begin{figure}[t]
    \centering
    \includegraphics[width=1.00\linewidth]{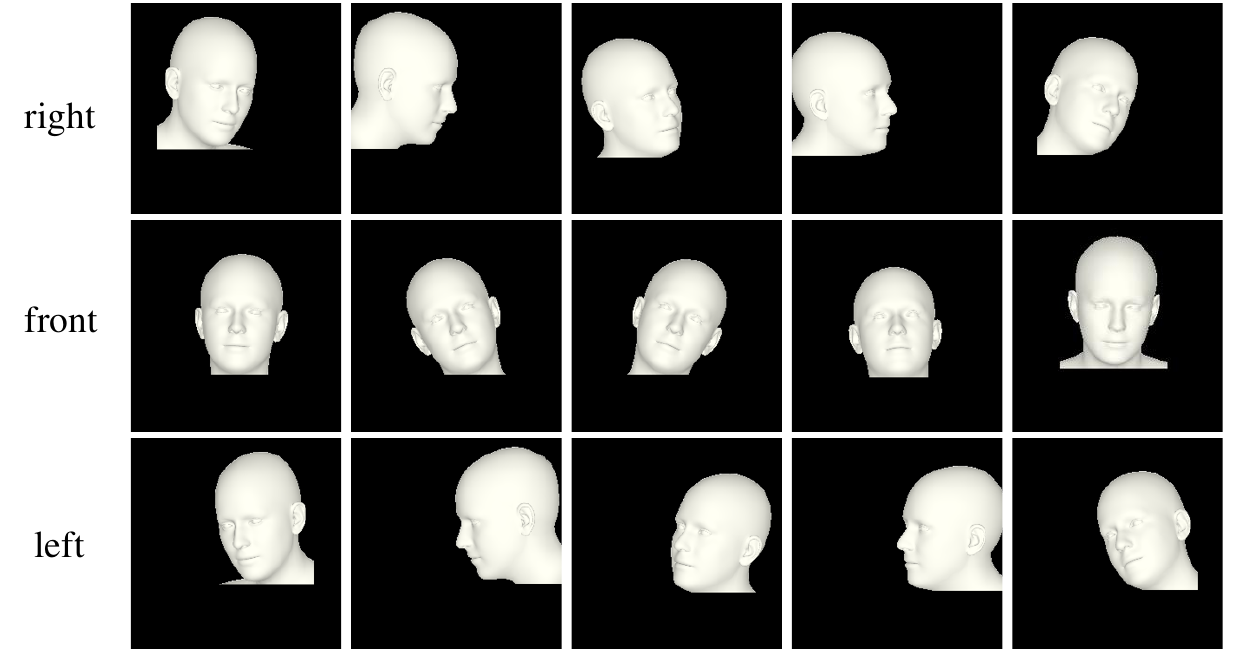}
    \caption{The prepared pose library. Each row corresponds to a class of pose respectively.}
    \label{fig:pose_lib}
\end{figure}

For the pose control $\mathbf{I}_p$, users can choose from a library of pre-defined poses or specify the pose by providing a target image.
The library consists of 15 template $\mathbf{I}_p$ with different poses, which are created by adjusting the mesh parameters.
The pose library is shown in \cref{fig:pose_lib}.
If the user provides a target image and requests a pose generation based on it, we use OSX to estimate the SMPL-X parameters and then construct $\mathbf{I}_p$ following the same procedure used during training.

Regarding the appearance control $\mathbf{I}_a$, users have the option to provide any face image with the desired appearance or omit the appearance branch to allow for free generation of the model face.
It is recommended to segment out any earrings present in the provided face image before using it as $\mathbf{I}_a$, as the earring content may interfere with the desired earring.

\begin{table*}[]
    \centering
    \begin{tabular}{ll|l|l}
    \hline
    \multicolumn{2}{l|}{Classes}                          & Standard                              & $\langle * \rangle$                \\ \hline
    \multicolumn{1}{l|}{\multirow{3}{*}{scale}} & small   & face height ratio $\sim$ [0, 0.4)            & `small face'       \\
    \multicolumn{1}{l|}{}                       & medium  & face height ratio $\sim$ [0.4, 0.7)          & `medium size face' \\
    \multicolumn{1}{l|}{}                       & big     & face height ratio $\sim$ [0.7, 1.0]          & `big face'         \\ \hline
    \multicolumn{1}{l|}{\multirow{3}{*}{pose}}  & right   & $\beta$ $\sim$ [$20^\circ$, $90^\circ$]   & `facing right'     \\
    \multicolumn{1}{l|}{}                       & front   & $\beta$ $\sim$ ($-20^\circ$, $20^\circ$)  & `facing front'     \\
    \multicolumn{1}{l|}{}                       & left    & $\beta$ $\sim$ [$-90^\circ$, $-20^\circ$] & `facing left'      \\ \hline
    \multicolumn{1}{l|}{race}                   & 6 classes & determined by \textit{deepface}~\cite{deepface}                & the race name      \\ \hline
    \end{tabular}
    \caption{The definition of all the classes of the scale, the pose and the race. $\beta$ denotes the relative angle between the norm vector of the model face and the y-z plane, and we ignore the cases that the model is facing back. The 6 classes of race are `asian', `indian', `black', `white', `middle eastern' and `latino hispanic'.}
    \label{tab:classes}
\end{table*}

\subsection{Experimental Details}
\subsubsection{Comparison with DreamBooth and Custom Diffusion}
To compare our method with tuning-based customized generative models~\cite{dreambooth}~\cite{custom-diffusion}, we follow the users' practice of using the pipeline for each earring.
Firstly, we manually place the earring image onto a black background as the conditioning image.
We randomly determine the position of the earring, ensuring that the positions are diverse and reasonable for advertising purposes.
Secondly, we randomly choose the face height ratio within the range of 0.3 to 0.7, which is the most common range in advertising images, and construct the scale map $\mathbf{I}_s$ as described in \cref{subsec:construct}.
Moreover, we randomly choose the pose control $\mathbf{I}_p$ from the library described in \cref{fig:pose_lib} to cover various poses.
We disable the appearance branch by setting the weighting factor $\alpha(t)$ to 0, allowing the pipeline to generate model faces freely.

\subsubsection{Comparison with Textual Control}
To validate the control effectiveness of our method, we compare our method with textual control in \cref{subsec:text}.
We define a total of 3 classes for face scales, 3 classes for poses, and 6 classes for races, as shown in \cref{tab:classes}.
Additionally, for each class, we design a short phrase $\langle * \rangle$ which is added to the default caption to serve as the textual control. 
We select 300 images from the 3000 test images to conduct the experiment.
The earrings in these selected images are segmented out as the conditioning image. 
We use the original Control-Net architecture to build the textual control counterparts.
Three models are trained for scale control, pose control, and race control respectively. 
Each model is used to generate 300 images, randomly selecting the control $\langle * \rangle$ from the corresponding control type.

For our method, we prepare 3 candidate scales, with face height ratios of 0.3, 0.5, and 0.8 respectively for each scale class.
We use the 3 groups of $\mathbf{I}_p$ depicted in \cref{fig:pose_lib} to represent the 3 classes of poses.
For scale control, we randomly construct $\mathbf{I}_s$ from the 3 candidates and use the target image to generate $\mathbf{I}_p$.
Conversely, for pose control, we randomly choose $\mathbf{I}_p$ from the 3 types and use the target image to generate $\mathbf{I}_s$. 
The appearance branch is dropped for both cases.
For race control, we use the target image to generate $\mathbf{I}_p$ and $\mathbf{I}_s$, and randomly select a face image of the corresponding race from the training data to serve as $\mathbf{I}_a$. 
After obtaining 300 generated images for scale control, pose control, and race control respectively, we examine the class label of the generated images and calculate the class alignment rate to measure the effectiveness of the control.

\subsection{User study of the effects of branch balancing}

In \cref{subsec:ablation}, we analyze the effects of the branch balancing schemes qualitatively and the examples are shown in \cref{fig:ablation}.
To make a quantitative comparison, we conduct a user study on our method, our method without TDW, our method without STD-Norm and TDW, and IP-Adapter.
We generate 1000 images for each method and recruit 20 volunteers to answer the question: "If this image is consistent with the text prompts?".
Each volunteer judges 50 images for each method.
The text prompts are randomly selected from a set of candidate prompts.
Each prompt starts with \textit{`A beautiful woman wears earrings, best quality,'} and ends with a phrase like \textit{`with glasses'}, \textit{`wears a hat'} and etc.
The results are shown in \cref{tab:ablation-user}.
It is clear to see both STD-Norm and TDW are essential for the alignment between the generated images and the text prompts.
Also, IP-Adapter fails to combine the earring image with a generated human face according to the text prompts, thus it can not be directly used for \textit{strictly-ID-preserved} earring advertising image generation.

\begin{table}[]
\centering
\caption{The quantitative evaluation of the branch balancing strategies.}
\begin{tabular}{l|c|c|c|c}
\hline
                  & Ours & w/o. TDW & w/o. STD-Norm \& TDW & IP-Adapter \\ \hline
Textual Alignment & \textbf{97.2\%}     & 65.3\%    & 8.2\% & 0.6\%     \\ \hline
\end{tabular}
\label{tab:ablation-user}
\end{table}

\end{document}